\documentclass{article}

\usepackage{amsmath}       % 解决 \text 报错
\usepackage{amssymb}
 \usepackage[preprint]{neurips_2025}

\usepackage[utf8]{inputenc} % allow utf-8 input
\usepackage[T1]{fontenc}    % use 8-bit T1 fonts
\usepackage{hyperref}       % hyperlinks
\usepackage{url}            % simple URL typesetting
\usepackage{booktabs}       % professional-quality tables
\usepackage{amsfonts}       % blackboard math symbols
\usepackage{nicefrac}       % compact symbols for 1/2, etc.
\usepackage{microtype}      % microtypography
\usepackage{xcolor}         % colors
\usepackage{booktabs}      % 提供 \bottomrule 支持

\usepackage{algorithm, multirow, booktabs, graphicx, wrapfig, subcaption}
\usepackage{booktabs}   % 提供 \toprule, \midrule, \bottomrule
\usepackage{multirow}   % 提供 \multirow
\usepackage{graphicx}   % 提供 \resizebox
\usepackage{makecell}   % 提供 \makecell
\usepackage{enumerate}
\usepackage{amsfonts}
\usepackage{amsthm}
\usepackage{newtxtt}
\usepackage{diagbox}
\usepackage{colortbl}
\usepackage{xspace}
\usepackage{adjustbox}
\usepackage{tabularx}
\usepackage{mathtools}
\usepackage{tikz}
\usepackage{enumitem}
\usepackage{silence}
\usepackage{dsfont}
\usepackage{xfakebold}
\usepackage{hepnames}
\usepackage{comment}
\usepackage{cuted}
\usepackage{url}
\usepackage{caption}
\usepackage{subcaption}

\usepackage[utf8]{inputenc} % allow utf-8 input
\usepackage[T1]{fontenc}    % use 8-bit T1 fonts
\usepackage{hyperref}       % hyperlinks
\usepackage{url}            % simple URL typesetting
\usepackage{booktabs}       % professional-quality tables
\usepackage{amsfonts}       % blackboard math symbols
\usepackage{nicefrac}       % compact symbols for 1/2, etc.
\usepackage{microtype}      % microtypography

\usepackage{algorithm}     % 算法外框
\usepackage{algorithmic}   % 必须是这个，不能是 algpseudocode

\usepackage{graphicx}
\usepackage{booktabs}
\usepackage{orcidlink}

\title{Geometric Autoencoder for Diffusion Models}

\author{%
  Hangyu Liu\textsuperscript{1}\quad Jianyong Wang\textsuperscript{2}\quad Yutao Sun\textsuperscript{1,2${\dag}$} \\
  $^{1}$Shanghai Innovation Institute \quad  $^{2}$Tsinghua University \\
}
\begin{document}

\maketitle

\begin{NoHyper}
\def\thefootnote{$\star$}\footnotetext{$^\dag$Corresponding author.}
\end{NoHyper}

\begin{abstract}
Latent diffusion models have established a new state-of-the-art in high-resolution visual generation. Integrating Vision Foundation Model priors improves generative efficiency, yet existing latent designs remain largely heuristic. These approaches often struggle to unify semantic discriminability, reconstruction fidelity, and latent compactness. In this paper, we propose Geometric Autoencoder (GAE), a principled framework that systematically addresses these challenges. By analyzing various alignment paradigms, GAE constructs an optimized low-dimensional semantic supervision target from VFMs to provide guidance for the autoencoder. Furthermore, we leverage latent normalization that replaces the restrictive KL-divergence of standard VAEs, enabling a more stable latent manifold specifically optimized for diffusion learning. To ensure robust reconstruction under high-intensity noise, GAE incorporates a dynamic noise sampling mechanism. Empirically, GAE achieves compelling performance on the ImageNet-1K $256 \times 256$ benchmark, reaching a gFID of 1.82 at only 80 epochs and 1.31 at 800 epochs without Classifier-Free Guidance, significantly surpassing existing state-of-the-art methods. Beyond generative quality, GAE establishes a superior equilibrium between compression, semantic depth and robust reconstruction stability. These results validate our design considerations, offering a promising paradigm for latent diffusion modeling. Code and models are publicly available
at~\url{https://github.com/sii-research/GAE}.
\end{abstract}

\section{Introduction}
\label{sec:intro}
Diffusion models~\citep{ho2020denoising,song2020score} have rapidly become the main component of visual generation tasks, offering a level of quality and flexibility of synthesis that has redefined the production of complex high-resolution media. Compared to direct pixel-level generation~\citep{li2025back,chen2025pixelflow}, latent diffusion~\citep{vahdat2021score,rombach2022high} that operate within a compressed space produced by a Variational Autoencoder (VAE) have proven significantly more effective. This paradigm now serves as the technical foundation for many current state-of-the-art generative models~\citep{esser2024scaling,seedream4,wan}.

Recently, the integration of semantic representations into diffusion models has emerged as a key strategy to improve learning efficiency and accelerate convergence. To this end, several parallel research trajectories have emerged to bridge the gap between reconstructive autoencoding and representation learning. First, VAE with semantic supervision~\citep{yao2025reconstruction,chen2025aligning} attempts to enhance the latent space by introducing supervised signals during the Autoencoder training stage. Second, alignment-based methods~\citep{yu2024representation,leng2025repa} utilize auxiliary losses to align VAE latents with the feature spaces of Vision Foundation Models (VFMs). Alternatively, another significant direction explores the direct integration of VFMs with reconstruction capability~\citep{zheng2025diffusion,gao2025one,bi2025vision,gui2025adapting}, to inherit powerful semantic priors for the generative process. Together, these methodologies demonstrate that unifying perceptual understanding with generative reconstruction is a critical frontier for developing more efficient and capable diffusion models.
\begin{figure}[t]
    \centering
    % --- 左子图：Semantic-Reconstruction Pareto Frontier ---
    \begin{subfigure}{0.49\linewidth}
        \centering
        \includegraphics[width=\linewidth]{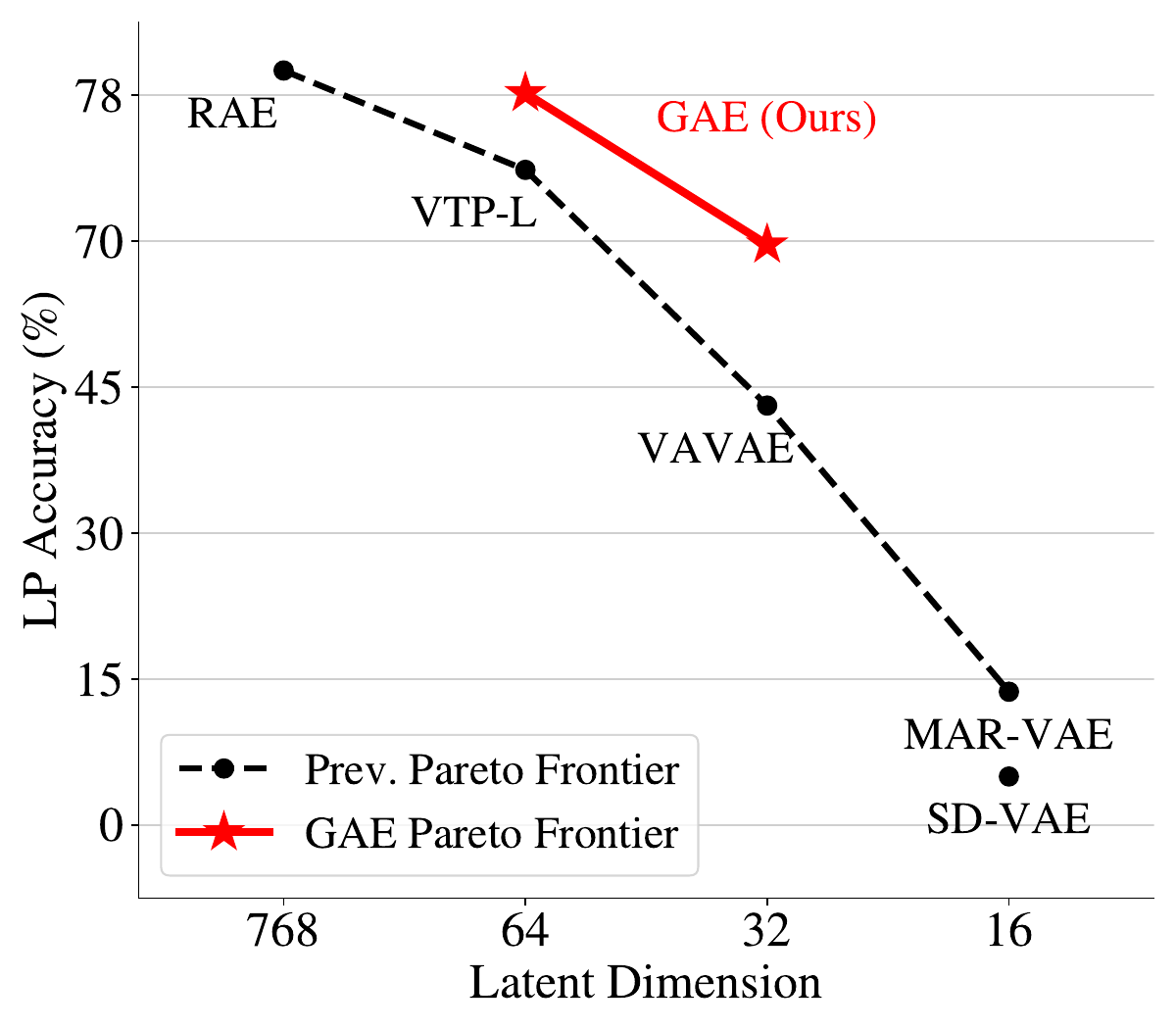}
        % \caption{} % 自动标注 (a)
        \label{fig:teaser_left}
    \end{subfigure}
    \hfill % 在两个子图之间填满空白
    % --- 右子图：Generative Convergence Efficiency ---
    \begin{subfigure}{0.49\linewidth}
        \centering
        \includegraphics[width=\linewidth]{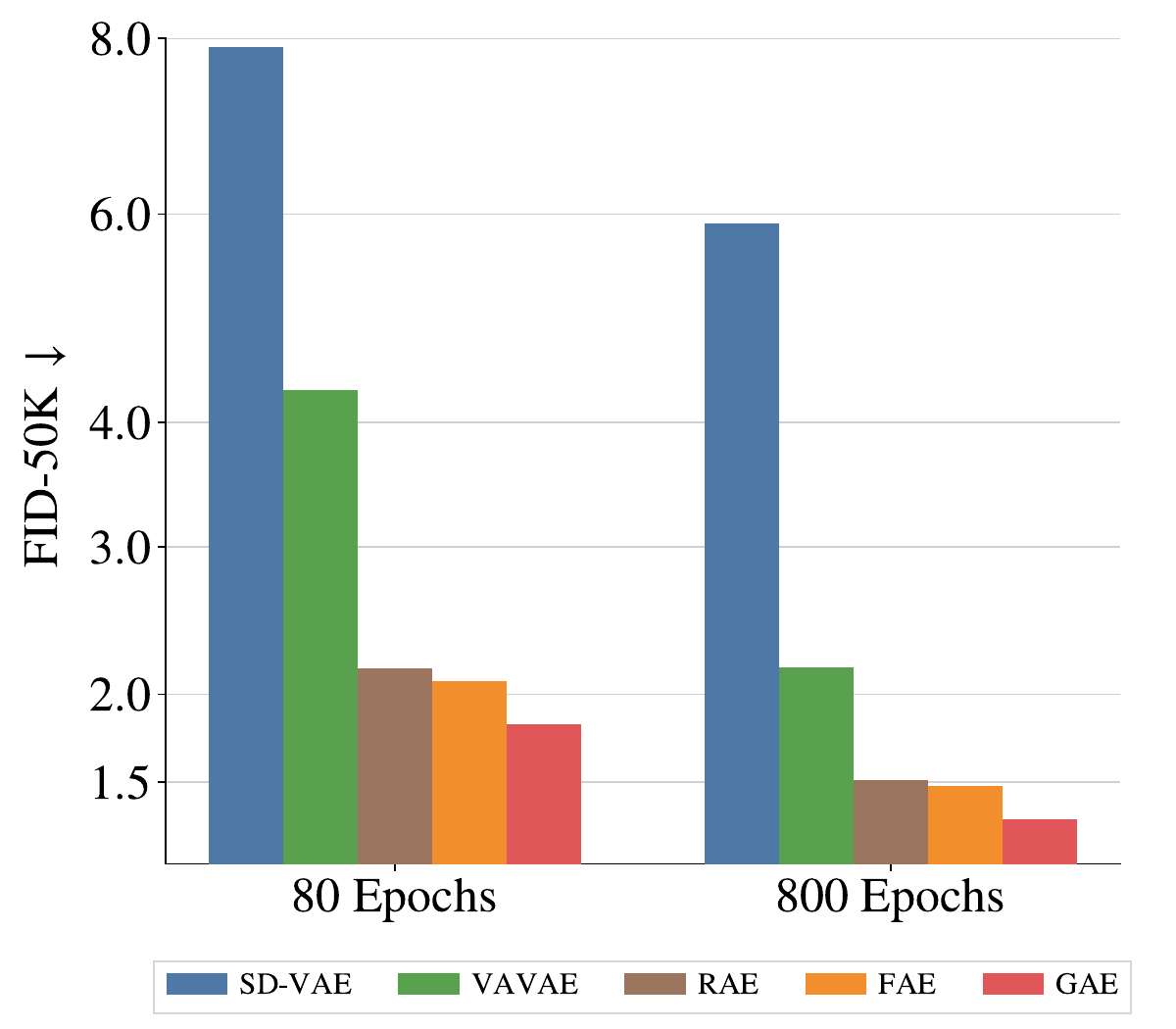}
        % \caption{} % 自动标注 (b)
        \label{fig:teaser_right}
    \end{subfigure}
    \caption{\textbf{Overview of GAE performance.} \textbf{Left:} GAE establishes a superior Pareto curve between linear probing accuracy and latent dimension. \textbf{Right:} GAE with 32 latent dimension significantly accelerates convergence and delivers superior generation results at both 80 and 800 training epochs.}
    \label{fig:trilemma}
\end{figure}

Despite these advancements, the design of latent spaces for diffusion models remains largely heuristic. This is due to a fundamental lack of principled guidance, leaving the complex interplay between various latent properties poorly understood.
First, alignment-based methods attempt to bridge the gap between reconstructive and semantic spaces, but the diverse range of alignment strategies frequently results in sub-optimal representations. Second, while lower dimensionality usually benefits diffusion training, VFM-based Autoencoder lacks the flexibility to adjust the latent space. Furthermore, despite addressing both semantic and dimensional requirements, frameworks like VTP~\citep{yao2025towards} still exhibit suboptimal generation quality due to inferior reconstruction stability. These discrepancies highlight that existing heuristic methods fail to analyze the diffusion representation space from a fundamental perspective, leading to sub-optimal generative performance.

In this paper, we propose \textit{Geometric Autoencoder} (GAE), a principled framework designed to systematically address the critical challenges in latent space design for diffusion models. First, by analyzing various alignment paradigms, we identify that constructing a low-dimensional semantic representation from existing VFMs provides the most effective guidance for the latent space of Autoencoder. Second, we leverage a latent normalization mechanism to optimize the distribution for diffusion learning, replacing the restrictive KL-divergence typically found in standard VAEs to provide a more stable and scalable latent manifold. Finally, by incorporating the dynamic noise sampling mechanism~\citep{sun2024multimodal}, GAE achieves robust reconstruction stability under high-intensity noise regimes. Collectively, these components allow GAE to overcome the limitations of prior heuristic designs, establishing a more robust and principled foundation for latent diffusion modeling.

Empirically, GAE achieves compelling performance in high-fidelity visual generation. On ImageNet~\citep{deng2009imagenet} $256\times256$ generation, GAE achieves a gFID of 1.82 at 80 epochs and further reaches 1.31 at 800 epochs without Classifier-Free Guidance, significantly surpassing existing state-of-the-art methods.
Beyond generative performance, our analysis demonstrates that GAE achieves a superior pareto frontier between compression and semantic information. Specifically, GAE attains impressive linear probing accuracies of 69.4\% and 78.3\% at 32 and 64 latent dimensions respectively, while maintaining robust reconstruction stability.
We further validate the effectiveness of our framework across various latent dimensions, with ablation studies confirming the critical contribution of each design component. These results substantiate our design principles, offering a more principled and effective roadmap for future advancements in latent diffusion modeling.

\section{Related Work}
\label{sec:related}
\paragraph{Vision Foundation Models.} 
The landscape of representation learning has been redefined by Vision Foundation Models (VFMs), which provide robust semantic priors through two primary paradigms. The first category comprises vision-language models such as CLIP~\citep{radford2021learning} and SigLIP~\citep{zhai2023sigmoid,tschannen2025siglip}, which align visual features with textual descriptions through large-scale contrastive learning. The second category focuses on visual self-supervised learning, ranging from reconstructive frameworks like MAE~\citep{he2022masked} to discriminative approaches such as SimCLR~\citep{chen2020simple} and DINOv2~\citep{oquab2023dinov2}. While these models excel at visual understanding, their architectures typically lack inherent data compression, and their representations provide no performance guarantees for generative reconstruction. Consequently, these feature spaces cannot be directly adopted as latent representations for diffusion models without reconciling the gap between semantic discriminability and reconstructive fidelity.

\paragraph{Representation Alignment for Generation.}
The integration of semantic representations into diffusion models has recently emerged as a pivotal strategy for enhancing learning efficiency and accelerating convergence. This research landscape is primarily characterized by three converging trajectories that seek to bridge the historical gap between reconstructive autoencoding and representation learning.
First, researchers explored the augmentation of VAE through semantic supervision, such as VA-VAE~\citep{yao2025reconstruction}, AlignTok~\citep{chen2025aligning}, which introduces supervised signals during the autoencoder training phase to enforce a more discriminative latent space. 
Second, alignment-based methodologies such as REPA~\citep{yu2024representation} and REPA-E~\citep{leng2025repa} utilize auxiliary objective functions to align the inner representation of latent diffusion model with the high-dimensional feature spaces of pre-trained Vision Foundation Models.
Recently, empowering existing Vision Foundation Models with reconstruction capabilities~\citep{zheng2025diffusion,gao2025one,bi2025vision,gui2025adapting} instead of training a dedicated autoencoder from scratch also demonstrates superior performance. Adapting the strong vision encoders allows the diffusion objective to directly inherit robust semantic priors.
Beyond the adaptation methods, VTP~\citep{yao2025towards} explores a holistic approach by joint-training foundational models with contrastive, self-supervised, and reconstruction objectives.
Despite these advancements, the current landscape remains fragmented as most methodologies are primarily heuristic and lack a systematic analysis, leaving a critical gap in the principled integration of perception and reconstruction for diffusion models.

\section{Geometric Autoencoder}
\label{sec:method}

In this section, we present the GAE framework. We first detail the overall architecture, followed by the formulation of the training objectives.

\begin{wrapfigure}[13]{r}{0.5\linewidth}
\vspace{-3.0em}
    \centering
    \includegraphics[width=\linewidth]{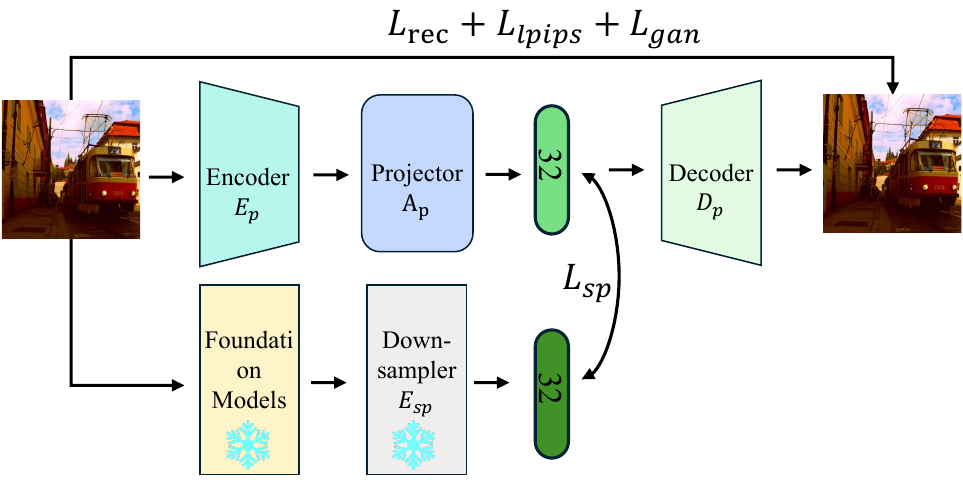}
    \caption{Overview of the GAE architecture. The input image is processed through a pixel-level branch ($E_p$, $A_p$) and a frozen semantic branch (VFM, $E_{sp}$). A Semantic Preservation loss $L_{sp}$ aligns the latent mean with the features from the decoupled Semantic Teacher.}
    \label{fig:main_arch}
\end{wrapfigure}

\subsection{Architecture}
\label{sec:method_architecture}

As illustrated in Fig.~\ref{fig:main_arch}, our framework employs a dual-branch design consisting of an encoder, including Encoder $E_p$, Projector $A_p$, Decoder $D_p$, and a frozen semantic teacher pipeline utilizing a VFM followed by Downsampler $E_{sp}$.

\paragraph{Transformer Autoencoder.}
We leverage the standard ViT~\citep{vit} backbone in both Encoder and Decoder with modern architectural refinements, including RMSNorm~\citep{zhang2019root} and the SwiGLU~\citep{glu}. ViT-based Autoencoder~\citep{magi,sun2024multimodal} achieves significantly higher throughput and better scalability during both training and inference compared with convolution-based neural network.

\paragraph{Latent Normalization.}
Given an input image $x$, the pixel encoder $E_p$ extracts spatial features, which are subsequently mapped to a compact latent space through the projector $A_p$. To ensure numerical stability and a well-distributed latent space~\citep{ke2025hyperspherical}, we apply a parameter-free regularization via RMSNorm. This operation projects the features onto a unit hypersphere, effectively bounding the latent values and preventing training collapse.

\paragraph{Dynamic Noise Sampling.}
Following the $\sigma$-VAE~\citep{sun2024multimodal} formulation, we introduce a dynamic sampling mechanism to enhance the robustness of the latent representation. Instead of a fixed variance, we sample a noise scale $\sigma$ and perturb the normalized latent mean $\mu$ with Gaussian noise $\boldsymbol{\epsilon}$. This stochastic process allows the model to learn a continuous manifold under varying noise levels. The latent vector $z$ is then fed into the pixel decoder $D_p$ to reconstruct the input $\hat{x}$. Formally, the reconstruction process is defined as:
\begin{equation}
\begin{aligned}
\mu &=\text{RMSNorm}(A_p(E_p(x))) \\
z &= \mu + |\sigma| \odot \boldsymbol{\epsilon}, \text{where} ~ \boldsymbol{\epsilon} \sim \mathcal{N}(0, 1), ~ \sigma \sim \mathcal{N}(0, C_{\sigma}) \\
\hat{x} &= D_p(z)
\nonumber
\end{aligned}
\end{equation}
where $C_{\sigma}$ is a hyperparameter controlling the noise sampling level.

\subsection{Objective}
\label{sec:method_objective}

\paragraph{Image Reconstruction.}
The reconstruction process is optimized using a multi-objective loss function comprising pixel-level $L_{1}$ loss ($L_{rec}$), perceptual loss ($L_{lpips}$), and adversarial loss ($L_{gan}$):
\begin{equation}
    \mathcal{L}_{pixel} = \lambda_{\text{rec}} \mathcal{L}_{\text{rec}} + \lambda_{\text{lpips}} \mathcal{L}_{\text{lpips}} + \lambda_{\text{gan}} \mathcal{L}_{\text{gan}} 
\end{equation}

\paragraph{Semantic Preservation.}
To ensure that the compressed latent space retains high-level semantic integrity and discriminability, we introduce a Semantic Preservation loss. Specifically, we align the latent mean $\mu$ produced by the pixel branch with the feature representations extracted from the Downsampler $E_{sp}$. The construction of $E_{sp}$ will be introduced in Sec.~\ref{sec:taxonomy}. The alignment is supervised via MSE loss:
\begin{equation}
    \mathcal{L}_{sp} = \| \mu - E_{sp}(f_{vfm}(x)) \|^2_2
\end{equation}

\paragraph{Removal of KL Objective.}
We remove the Kullback-Leibler divergence penalty typically required in VAE frameworks~\citep{kingma2013auto}. Traditional VAEs rely on KL divergence to push the latent distribution toward a standard Gaussian to facilitate sampling. However, by employing latent normalization, we enforce a hard geometric constraint that projects the latent mean $\mu$ onto a unit hypersphere, ensuring it remains bounded and well-distributed. Coupled with dynamic noise sampling mechanism, this approach enhances reconstruction robustness without the instability of a weighted KL term. This design preserves the structural integrity of the latent space, which is more conducive to the subsequent denoising process in diffusion training.

\paragraph{Overall Objective.} Integrating the previously defined components, the total optimization objective is computed as:

\begin{equation}
    \mathcal{L}_{total} = \lambda_{\text{rec}} \mathcal{L}_{\text{rec}} + \lambda_{\text{lpips}} \mathcal{L}_{\text{lpips}} + \lambda_{\text{gan}} \mathcal{L}_{\text{gan}} + \lambda_{\text{sp}} \mathcal{L}_{\text{sp}}
\end{equation}

Following standard practices in generative modeling~\citep{rombach2022high,ke2025hyperspherical}, we empirically set the balancing hyperparameters to $\lambda_{\text{rec}} = 1.0$, $\lambda_{\text{lpips}} = 1.0$, and $\lambda_{\text{gan}} = 0.5$. For the semantic preservation term, we set $\lambda_{\text{sp}} = 1.0$ to ensure a strong alignment between the latent space and the teacher’s semantic representations.

\section{Latent Alignment for Antoencoder}
\label{sec:taxonomy}
In this section, we systematically investigate the design space of semantic alignment between generative autoencoders and Vision Foundation Models. We argue that the stage at which semantic supervision is introduced significantly impacts both reconstruction fidelity and semantic discriminability. Through a series of pilot studies, we identify Latent Alignment, a direct supervision at the compact bottleneck as the most effective paradigm for preserving semantic integrity. To implement this, we further propose a parametric downsampler to bridge the dimensionality gap, ensuring that the autoencoder inherits the rich, discriminative priors of the VFM without compromising generative performance.

\begin{figure}[!t]
  \centering
  \includegraphics[width=1.0\linewidth]{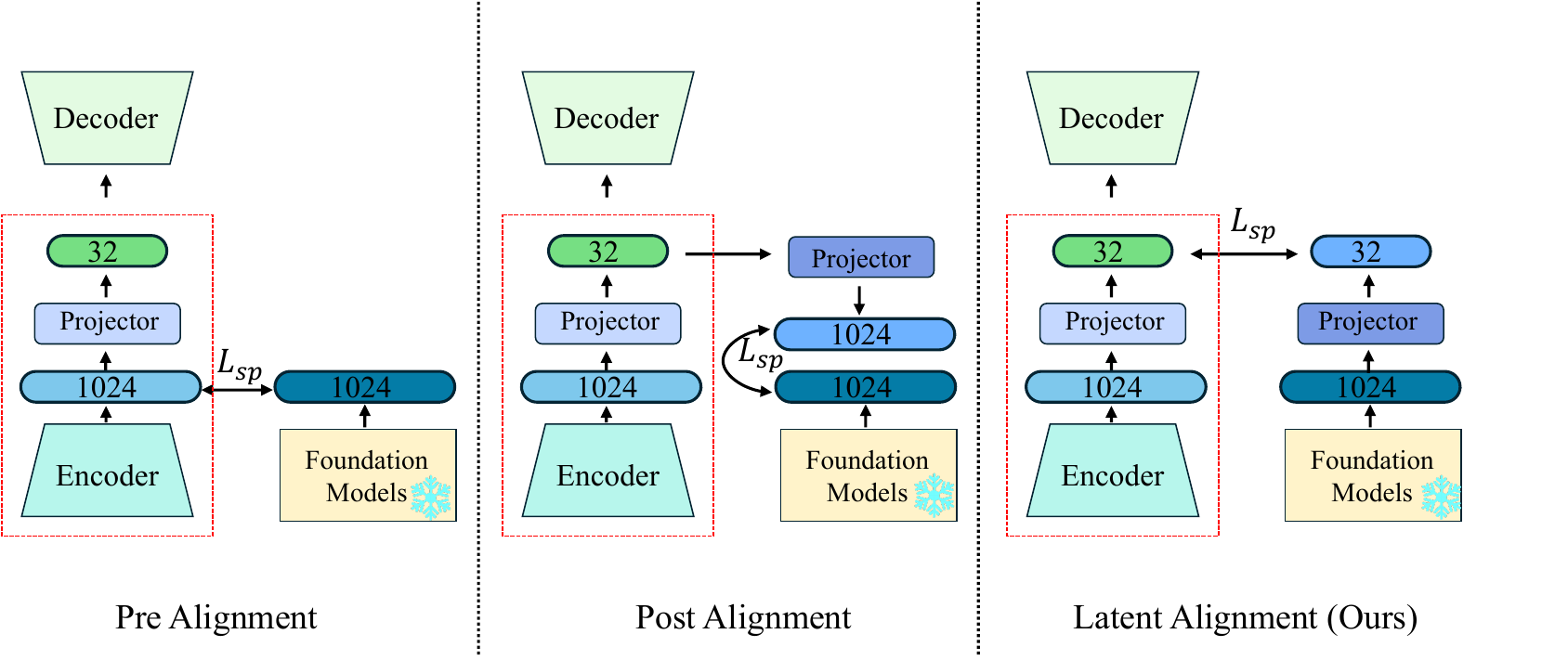} 
  \caption{Illustration of the three latent alignment paradigms: The term $L_{sp}$ aligns the AE representations with those of the VFM. \emph{Pre Alignment} aligns high-dimensional encoder features directly; \emph{Post Alignment} projects AE latents back to a high-dimensional space for supervision; \emph{Latent Alignment} operates within the compressed latent bottleneck via a projection of VFM features.
  }
  \label{fig:alignment_taxonomy}
  \vspace{-1.5em}
\end{figure}

\subsection{Design Spaces for Semantic Alignment}
\label{sec:design_sp}

Fundamental challenge in integrating generative autoencoders with Vision Foundation Models lies in their inherent dimensionality mismatch. While pixel-level Autoencoder typically compresses images into a highly compact latent space (e.g., $d=32$) to facilitate efficient diffusion training, VFMs often operate in a high-dimensional semantic space (e.g., $d=1024$ for ViT-L). The specific stage of alignment dictates the interplay between semantic density and reconstruction fidelity. We categorize and evaluate three distinct alignment methodologies to systematically explore this design space.

\paragraph{Taxonomy of Alignment.}
We categorize existing semantic alignment methodologies into three distinct paradigms based on the architectural stage at which dimensionality reconciliation occurs:
\begin{itemize}
    \item \textbf{Pre Alignment}: This strategy aligns the encoder's intermediate high dimensional features before the compressed latent with the VFM prior.
    \item \textbf{Post Alignment}: This paradigm projects the compressed latents back into a high-dimensional space via an additional expansion layer for high dimensional supervision.
    \item \textbf{Latent Alignment (Ours)}: This paradigm reconciles the dimensionality mismatch by introducing a semantic downsampler to project the VFM's high-dimensional outputs into a compact space that directly matches the AE's latent dimensions.
\end{itemize}

\paragraph{Pilot Study Setup.}
We instantiate the framework using ViT-L backbones for the Autoencoder and a frozen DINOv2-L/14 as the semantic teacher. For Latent Alignment, we utilize a pre-computed SVD projection matrix derived from $64,000$ ImageNet samples to map the $1024$-dim VFM features onto the $32$-dim manifold. All variants undergo a $60$-epoch training phase. We evaluate the learned representations across two dimensions: reconstruction fidelity and semantic discriminability via Linear Probing (LP) on the stand-alone latents.

\paragraph{Analysis and Findings.}
As detailed in Tab.~\ref{tab:pre_mid_post}, our analysis reveals a critical bottleneck in information flow. While Pre Alignment achieves a slight advantage in reconstruction, it suffers from a dramatic loss in semantic discriminability. This suggests that supervising the high-dimensional features prior to the projector does not guarantee that semantic integrity is maintained through the subsequent compression stage. In contrast, Latent Alignment demonstrates a remarkable capacity to inherit VFM priors, achieving the highest LP accuracy. This result is particularly significant given the simplicity of the SVD-based downsampler, which suggests that direct supervision at the bottleneck is the most effective way to anchor the latent space. 

% \begin{table}[t]
% \centering
% \caption{Quantitative comparison of alignment paradigms on ImageNet.}
% \label{tab:results}
% \begin{tabular}{l|c|cccc|c}
% \hline
% \textbf{Method} & \textbf{Dim} & \textbf{rFID} $\downarrow$ & \textbf{PSNR} $\uparrow$ & \textbf{LPIPS} $\downarrow$ & \textbf{SSIM} $\uparrow$ & \textbf{LP} $\uparrow$ \\ \hline
% Pre              & 32           & \textbf{0.40}           & \textbf{28.02}          & \textbf{0.101}           & \textbf{0.794}          & 20.9                                         \\
% Post             & 32           & 0.48                    & 26.53                   & 0.122                    & 0.744                   & 60.8                                         \\
% Mid  & 32           & 0.51                    & 25.68                  & 0.136                    & 0.714                   & \textbf{63.2}                       \\ \hline
% Teacher (SVD)    & 32           & -                        & -                        & -                        & -                       & 60.9                                         \\
% Teacher (DinoL)  & 1024         & -                        & -                        & -                        & -                                             & 83.7                   \\ \hline
% \end{tabular}

% \end{table}

\begin{table}[t]
\centering
\vspace{-1.0em}
\caption{Pilot Study of alignment paradigms on ImageNet.}
\label{tab:pre_mid_post}
\setlength{\tabcolsep}{10pt} % 优化间距
\begin{tabular}{l ccc c c} % 去掉了所有竖线
\toprule
\multirow{2}{*}{Metric} & \multicolumn{3}{c}{Autoencoder} & \multicolumn{2}{c}{Teacher} \\ 
\cmidrule(r){2-4} \cmidrule(l){5-6} % 使用 cmidrule 区分逻辑组
 & Pre & Post & Latent & SVD & DinoL \\ 
\midrule
rFID $\downarrow$ & \textbf{0.40} & 0.48 & 0.51 & -- & -- \\
LP $\uparrow$ & 20.9 & 60.8 & \textbf{63.2} & 60.9 & 83.7 \\ 
\bottomrule
\end{tabular}
\vspace{-1.0em}
\end{table}

\subsection{Training Regime of Semantic Teacher}
\label{sec:method_teacher}

While the non-parametric SVD projection provides a baseline for Latent Alignment, its token-wise operation neglects essential spatial correlations, thereby restricting the model's ability to construct a semantically coherent compressed space. To overcome this, we advocate for a parametric downsampling approach, where a neural network is optimized to distill high-dimensional VFM priors into a compact latent form. By learning a non-linear mapping instead of a static projection, the model can better reconcile structural integrity with latent compactness, ensuring that the resulting teacher representation is both efficient and discriminative.

\begin{wraptable}[8]{r}{0.35\textwidth} % 缩小宽度至 0.35\textwidth 更适合环绕
\centering
\vspace{-1.5em} % 根据需要微调上方间距
\caption{Ablation of Downsampler in Semantic Teacher.}
\label{tab:sp_ablation}
\begin{tabular}{lc}
\toprule
Arch & LP $\uparrow$ \\ \midrule
Single Attn           & 62.8                   \\
Attn + Linear         & 63.4                   \\
Attn + Patch Conv     & \textbf{75.6}          \\ \bottomrule
\end{tabular}
% \vspace{-7em} % 根据需要微调上方间距
\end{wraptable}

\paragraph{Training Objective.}
We leverage Feature Autoencoder~\citep{gao2025one} for the Semantic Teacher training. The framework consists of a frozen VFM backbone $f_{\text{vfm}}$, a parametric downsampler $E_{\text{sp}}$, and a lightweight 4-layer Llama-style~\citep{touvron2023llama} Transformer decoder $D_{\text{sp}}$. To ensure the compressed latents faithfully preserve the teacher's foundational knowledge, we pre-train the encoder-decoder pair using a feature-based cosine distillation objective. Specifically, we treat the intermediate patch tokens from the frozen VFM as semantic anchors and train the bottleneck to retain sufficient information for their directional recovery. For an input image $x$, the training loss $\mathcal{L}_{\text{spt}}$ is defined as the cosine distance between the teacher features and the reconstructed counterparts:
\begin{equation}
    \mathcal{L}_{\text{spt}} =  - \cos(D_{\text{sp}}(E_{\text{sp}}(f_{\text{vfm}}(x))), f_{\text{vfm}}(x))
\end{equation}
This ensures the resulting low dimensional latents capture the most discriminative semantic components. After this pre-training phase, $D_{\text{sp}}$ is discarded, and the frozen $E_{\text{sp}}$ serves as the finalized downsampler to provide semantic supervision for the autoencoder.

\paragraph{Downsampler Architecture.}
To determine the optimal configuration for semantic compression, we evaluate three downsampler architectures: (1) \textit{Single Attention} following FAE~\citep{gao2025one}; (2) \textit{Attention+Linear}, which employs a global linear projection following the attention module; and (3) \textit{Attention+Patch Conv}, which utilizes a non-overlapping patch convolution to project tokens into the latent space. As shown in Tab.~\ref{tab:sp_ablation}, we assess the semantic quality of the resulting embeddings via Linear Probing. The results demonstrate that our Patch Conv design significantly outperforms the baselines, confirming that a spatial-aware, end-to-end learned downsampler is essential for constructing a discriminative teacher latent space.

\section{Experiments}
\label{sec:experiments}

In this section, we provide the implementation details and evaluate the proposed framework on the ImageNet-1K~\citep{deng2009imagenet} benchmark.

\paragraph{Experimental Setup for Main Generation (Secs~\ref{sec:exp_class_conditional}, \ref{sec:scale_large_dim}).}
For our primary generation tasks, we instantiate GAE with a ViT-L backbone. We leverage DINOv2-L~\citep{oquab2023dinov2} as the frozen Visual Foundation Models. The Autoencoder is trained for 200 epochs with a global batch size of 1024 using the AdamW optimizer. For the generative model, we strictly follow the optimization strategy of LightningDiT-XL~\citep{yao2025reconstruction}, utilizing a constant learning rate of $2.0 \times 10^{-4}$, a batch size of 1024, and an EMA weight of 0.9999. To ensure stability, we apply QK-Norm for the 800-epoch runs, while the 80-epoch benchmarks are conducted without QK-Norm. Comprehensive details are provided in Appendix~\ref{sec:supp_hyp_parameters}.

\paragraph{Ablation Study Configuration.}
Unless otherwise specified, our ablation experiments adopt the ViT-L backbone for the GAE, with default hyperparameters set to $\lambda_{sp}=1.0$, $d=32$, and $C_\sigma=0.1$. These Autoencoders undergo a 100-epoch training phase, adhering to the identical learning rate schedule as the 200-epoch primary configuration. For generative assessment, we report results using a LightningDiT-XL backbone trained for 80-epoch without QK-Norm. Comprehensive details are provided in Appendix~\ref{sec:supp_ablation_configs}.

\paragraph{Evaluation Metrics.}
To provide a comprehensive assessment, we evaluate the proposed alignment paradigm across three primary dimensions:
\begin{itemize}
    \item Semantic Discriminability: We assess the semantic density of the learned latent space via Linear Probing accuracy. Following the protocol in Appendix~\ref{sec:supp_lp_settings}, we train a linear classifier on the frozen latents of ImageNet to evaluate how effectively the VFM's discriminative priors are preserved after compression.
    \item Reconstruction Fidelity: To ensure the autoencoder maintains high-quality image recovery, we report standard pixel-level and perceptual metrics, including PSNR, SSIM, and LPIPS, and rFID to measure the statistical distance between the original and reconstructed validation sets, capturing structural and textural consistency.
    \item Generation Quality: During generation stage, we evaluate the performance of Latent Diffusion with gFID, Inception Score, Precision and Recall to assess image quality, diversity, fidelity and sample coverage.
\end{itemize}

\begin{table*}[t]
\centering
\vspace{-2.0em}
\caption{\label{tab:main_results} Quantitative comparison of class-conditional Image Generation on ImageNet 256$\times$256. GAE achieves 1.31 FID without guidance and 1.13 with Classifier-Free Guidance, outperforming all prior methods by a large margin. $^*$ denotes RAE utilizes the AutoGuidance~\citep{karras2024guiding} instead of standard Classifier-Free Guidance.}
\resizebox{\textwidth}{!}{%
\begin{tabular}{l|c|c|cccc|cccc}
\toprule
\textbf{Method} &
\textbf{Epochs} &
\textbf{\#Params} &
\multicolumn{4}{c|}{\textbf{Generation w/o CFG}} &
\multicolumn{4}{c}{\textbf{Generation w/ CFG}} \\
\cmidrule(lr){4-7}\cmidrule(lr){8-11}
 & 
 & 
 & \textbf{gFID} & \textbf{IS} & \textbf{Pre.} & \textbf{Rec.} 
 & \textbf{gFID} & \textbf{IS} & \textbf{Pre.} & \textbf{Rec.} \\
\midrule

\multicolumn{11}{c}{\textit{\textbf{AutoRegressive}}} \\
\midrule
MaskGIT~\citep{chang2022maskgit}     & 555  & 227M & 6.18   & 182.1 & 0.80 & 0.51 & --   & --   & --   & --   \\
LlamaGen~\citep{sun2024autoregressive}   & 300  & 3.1B & 9.38 & 112.9 & 0.69 & 0.67 & 2.18 & 263.3 & 0.81 & 0.58 \\
VAR~\citep{tian2024visual}             & 350  & 2.0B & --   & --    & --   & --   & 1.80  & 365.4 & 0.83 & 0.57 \\
MagViT-v2~\citep{yu2023language}  & 1080 & 307M & 3.65  & 200.5 & --   & --   & 1.78  & 319.4 & --   & --   \\
MAR~\citep{li2024autoregressive}             & 800  & 945M & 2.35  & 227.8 & 0.79 & 0.62 & 1.55 & 303.7 & 0.81 & 0.62 \\
\midrule

\multicolumn{11}{c}{\textit{\textbf{Latent Diffusion}}} \\
\midrule
MaskDiT~\citep{zheng2023fast}     & 1600 & 675M & 5.69 & 177.9 & 0.74 & 0.60 & 2.28 & 276.6 & 0.80 & 0.61 \\
DiT~\citep{peebles2023scalable}             & 1400 & 675M & 9.62  & 121.5 & 0.67 & 0.67 & 2.27 & 278.2 & \textbf{0.83} & 0.57 \\
SiT~\citep{ma2024sit}             & 1400 & 675M & 8.61  & 131.7 & 0.68 & 0.67 & 2.06 & 270.3 & 0.82 & 0.59 \\
FasterDiT~\citep{yao2024fasterdit} & 400  & 675M & 7.91 & 131.3 & 0.67 & \textbf{0.69} & 2.03 & 264.0 & 0.81 & 0.60 \\
MDT~\citep{gao2023masked}             & 1300 & 675M & 6.23  & 143.0 & 0.71 & 0.65 & 1.79 & 283.0 & 0.81 & 0.61 \\
MDTv2~\citep{gao2023mdtv2}         & 1080 & 675M & --     & --    & --   & --   & 1.58 & \textbf{314.7} & 0.79 & 0.65 \\
REPA~\citep{yu2024representation}           & 800  & 675M & 5.90   & --    & --   & --   & 1.42 & 305.7 & 0.80 & 0.65 \\

VA-VAE~\citep{yao2025reconstruction} & 64  & 675M &
5.14 & 130.2 & 0.76 & 0.62 &
2.11 & 252.3 & 0.81 & 0.58 \\
 & 800 & 675M &
2.17 & 205.6 & 0.77 & 0.65 &
1.35 & 295.3 & 0.79 & 0.65 \\

REPA-E~\citep{leng2025repa} & 80 & 675M & 3.46 & 159.8 & 0.77 & 0.63 & 1.67 & 266.3 & 0.80 & 0.63 \\
 & 800 & 675M & 1.70 & 217.3 & 0.77 & 0.66 & 1.15 & 304.0 & 0.79 & 0.66 \\

Aligntok~\citep{chen2025aligning} & 800 & 675M & 2.04 & 206.2 & 0.76 & 0.67 & 1.37 & 293.6 & 0.79 & 0.65 \\

RAE (DiT-XL)~\citep{zheng2025diffusion} & 800 & 676M &
1.87 & 209.7 & 0.80 & 0.63 &
1.41 & 309.4 & 0.80 & 0.63 \\

RAE (DiTDH-XL)~\citep{zheng2025diffusion}  & 80 & 839M &
2.16 & 214.8 & \textbf{0.82} & 0.59 &
-- & -- & -- & -- \\

 & 800 & 839M &
1.51 & 242.9 & 0.79 & 0.63 &
1.13$^*$ & 262.6$^*$ & 0.78$^*$ & 0.67$^*$ \\

FAE~\citep{gao2025one}
 & 80 & 675M &
2.08	& 207.6 &	\textbf{0.82}	& 0.59 &
1.70	& 243.8	& 0.82	& 0.61 \\
 & 800 & 675M &
1.48 & 239.8 &	0.81	& 0.63 &
1.29 & 268.0 & 0.80 & 0.64 \\

\midrule
\rowcolor{blue!8}
\textbf{GAE}
 & 80 & 675M &
1.82	& 220.4 &	\textbf{0.82}	& 0.61 &
1.48	& 265.2	& 0.80	& 0.62 \\
 
\rowcolor{blue!8}
 & 800 & 675M &
\textbf{1.31} & \textbf{254.4} &	0.80	& 0.64 &
\textbf{1.13} & 294.9 & 0.79 & \textbf{0.67} \\

\bottomrule
\end{tabular}%
}
\vspace{-1.0em}
\end{table*}

 \subsection{System-level Comparison}
\label{sec:exp_class_conditional}

In this section, we evaluate the class-conditional image generation performance of GAE on the ImageNet benchmark. As summarized in Tab.~\ref{tab:main_results}, GAE establishes new SOTA results across multiple generative metrics while maintaining an extremely compact 32-dimensional latent space. Complementary analyses regarding reconstruction performance are provided in Appendix~\ref{sec:supp_reconstruction}.

\paragraph{Generative Performance.} 
 When trained for 800 epochs, our framework attains a gFID of 1.31 without CFG and 1.13 with CFG. These results consistently surpass strong baselines such as FAE, which yields 1.48 and 1.29 respectively, and RAE, which achieves 1.51 and 1.13. Notably, while GAE matches the 1.13 gFID of RAE under guided sampling, RAE benefits from the more sophisticated AutoGuidance~\citep{karras2024guiding} protocol and larger model size. In contrast, our framework achieves comparable performance using only the standard CFG.

\paragraph{Training Efficiency and Convergence.} 
One of the most striking advantages of GAE is its remarkable training efficiency. Specifically, at only 80 epochs, GAE achieves a gFID of 1.82, which already surpasses the performance of VA-VAE trained for a full 800 epochs. This rapid convergence validates our core hypothesis that a semantically aligned latent space simplifies the learning objective for the subsequent diffusion model.

\paragraph{Semantic-Reconstruction Trade-off.} 
Despite its focus on semantic alignment, GAE does not sacrifice visual fidelity. As illustrated in the Pareto Frontier (Fig.~\ref{fig:trilemma}, left), GAE achieves a high LP accuracy of 69.4\%, which is substantially higher than the 43.1\% achieved by VA-VAE at the same latent dimension.

\begin{figure}[!t]
  \centering
  \vspace{-2.0em}
  \includegraphics[width=1.0\linewidth]{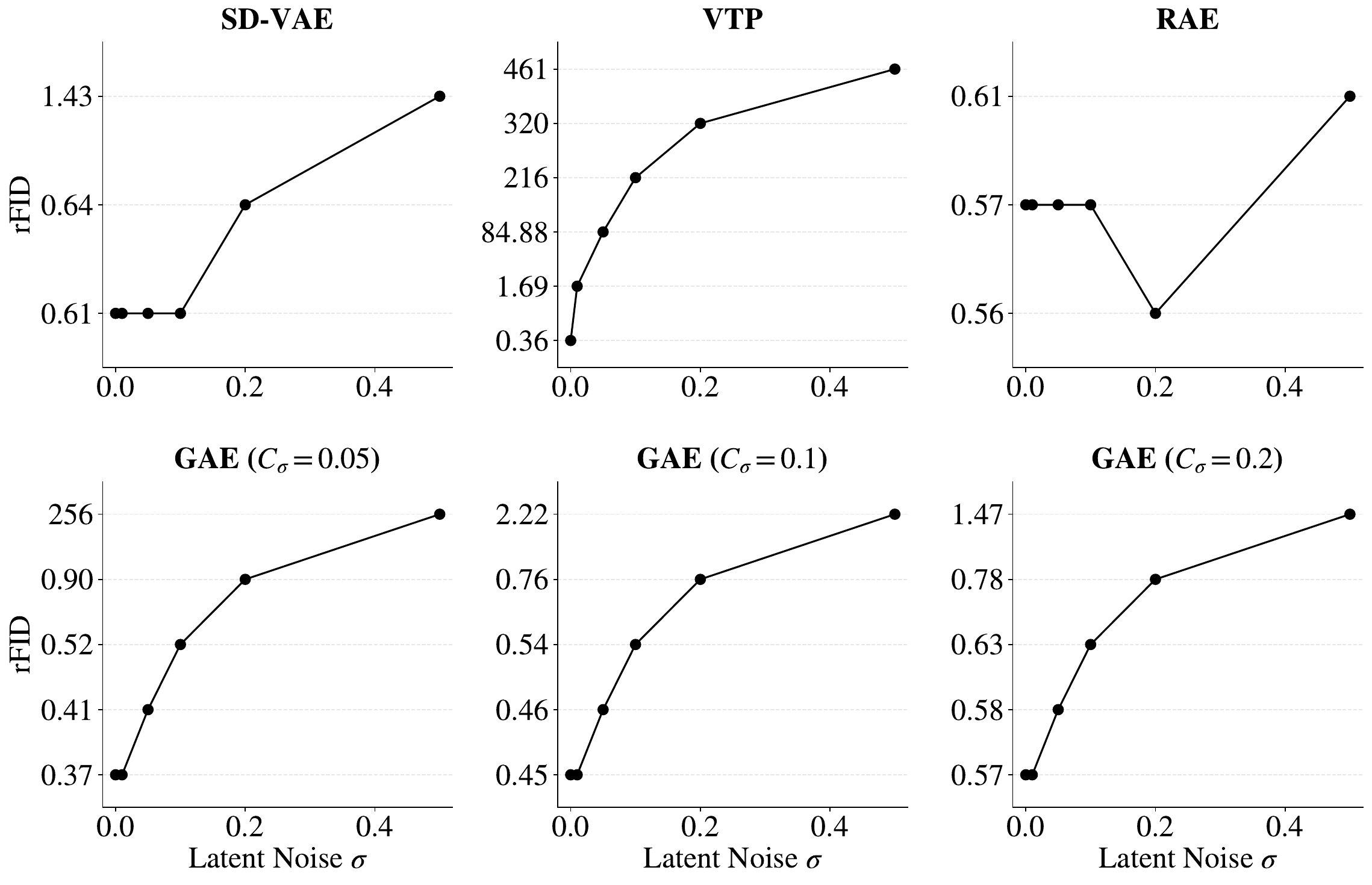} 
  \caption{Decoder stability against latent noise injection. We evaluate rFID by adding varying levels of Gaussian noise to the latent representations before decoding. The results demonstrate that models trained with higher $\sigma$ exhibit superior tolerance to latent distribution shifts, ensuring stable performance during generative sampling.}
  \label{fig:sigma}
  \vspace{-1.0em}
\end{figure}

\subsection{Reconstruction Stability}
\label{sec:exp_sigma}

\paragraph{Dynamic Noise Sampling.}
As illustrated in Fig.~\ref{fig:sigma}, our analysis reveals a fundamental trade-off between reconstruction fidelity and generative performance governed by the noise strength $C_\sigma$. While minimizing $C_\sigma$ inherently yields superior rFID by reducing the perturbation of the latent manifold, it may fail to provide robust reconstruction performance as the noise level increases. Specifically, for the $d=32$ configuration, although $C_\sigma=0.1$ maintains a slight lead in rFID during early training, this gap diminishes as the model scales, whereas $C_\sigma=0.2$ provides the necessary robustness to stabilize subsequent diffusion learning. In the $d=64$ case, we observe that generative performance effectively plateaus between $C_\sigma=0.2$ and $C_\sigma=0.3$. Consequently, we identify $C_\sigma=0.2$ for both $d=32$ and $d=64$ as the balanced configurations that harmonize semantic density with high-fidelity recovery, consistent with the experiment results in $\sigma$-VAE~\citep{sun2024multimodal}.

\paragraph{Analysis on Other Variants.}
We further extend our analysis to other autoencoder variants in Fig.~\ref{fig:sigma}. RAE demonstrates exceptional reconstruction resilience, where its rFID remains remarkably stable even at a high perturbation level of $\sigma=0.5$, which directly contributes to its superior gFID. In contrast, VTP exhibits a different behavior. While it achieves high semantic discriminability due to its training regime, its reconstruction performance is significantly more sensitive to latent perturbations. The sharp degradation in rFID hinders its generative performance, leading to sub-optimal gFID results.

\begin{table}[t]
\centering
\vspace{-2.0em}
\caption{Impact of standard deviation $C_\sigma$ and latent dimension $d$ on ImageNet-1K. gFID is reported without CFG. Autoencoders are trained on ImageNet-1K for 100 epochs.}
\label{tab:sigma_effect}
\begin{tabular}{c|c|cccc|c}
\toprule
Dim & $C_\sigma$ & rFID $\downarrow$ & PSNR $\uparrow$ & LPIPS $\downarrow$ & SSIM $\uparrow$ & gFID $\downarrow$ \\ \midrule
             & 0.05              & \textbf{0.37}            & \textbf{27.78}          & \textbf{0.101}           & \textbf{0.784}          & 2.58                      \\
$d=32$       & 0.1               & 0.45                     & 27.37                   & 0.107                    & 0.774                   & 2.36                      \\
             & 0.2               & 0.57                     & 26.99                    & 0.116                    & 0.759                   & \textbf{2.32}             \\ \midrule
             & 0.1               & \textbf{0.31}            & \textbf{29.60}          & \textbf{0.076}           & \textbf{0.845}          & 3.15                      \\
$d=64$       & 0.2               & 0.43                     & 28.92                   & 0.086                    & 0.827                   & 3.12                      \\
             & 0.3               & 0.49                     & 28.30                   & 0.094                    & 0.811                   & \textbf{3.11}             \\ \bottomrule
\end{tabular}
\vspace{-1.0em}
\end{table}

\subsection{Scalability to Higher Dimension}
\label{sec:scale_large_dim}

In this section, we investigate the scalability of the GAE framework by extending the latent space to 64 dimensions. We are using our GAE variant against two prominent 64-dimensional baselines, VTP-L and the 64-dim version of FAE. The results are summarized in Tab.~\ref{tab:large_dim}.

\paragraph{Semantic and Reconstruction.}
As shown in Tab.~\ref{tab:large_dim}, GAE demonstrates superior performance even as the latent dimensionality increases. In terms of semantic discriminability, GAE achieves the highest LP accuracy of 78.3\%, notably outperforming VTP-L (73.9\%). Regarding reconstruction fidelity, GAE maintains a substantial lead over FAE, with an rFID of 0.382 compared to 0.660.

\paragraph{Generation Performance.}
The generative advantages of GAE are also evident in the high-dimensional setting. Without the use of CFG, GAE achieves a gFID of 1.96 at only 80 training epochs, far surpassing VTP-L's 2.81. When training progresses to 160 epochs, GAE's gFID further improves to 1.53, which is considerably better than the 2.02 achieved by FAE at the same stage. These results confirm that our semantic alignment mechanism remains highly effective at larger scales, providing a superior foundation for diffusion-based image synthesis.

% \begin{table}[htbp]
%   \centering
%   \caption{Quantitative comparison in the 64-dimensional latent space on ImageNet $256\times256$. GAE-64 consistently outperforms existing 64-dim baselines in semantic alignment, reconstruction quality, and generative convergence speed.}
%   \label{tab:large_dim}
%   \resizebox{\linewidth}{!}{
%   \begin{tabular}{lccccccccc}
%     \toprule
%     \multirow{2}{*}{Method} & \multirow{2}{*}{dim} & \multirow{2}{*}{LP $\uparrow$} & \multicolumn{4}{c}{Reconstruction} & \multicolumn{3}{c}{gFID (w/o CFG) $\downarrow$} \\
%     \cmidrule(lr){4-7} \cmidrule(lr){8-9}
%     & & & rFID $\downarrow$ & PSNR $\uparrow$ & LPIPS $\downarrow$ & SSIM $\uparrow$ & 80 ep & 160 ep & 800ep \\
%     \midrule
%     VTP-L & 64 & 73.9 & 0.360 & 26.83 & 0.126 & 0.750 & 2.81 & -- & -- \\
%     FAE   & 64 & --   & 0.660 & --    & --    & --    & --   & 2.02 & -- \\
%     \midrule
%     \textbf{GAE} & \textbf{64} & \textbf{78.3} & \textbf{0.382} & \textbf{29.09} & \textbf{0.081} & \textbf{0.834} & \textbf{1.96} & \textbf{1.53} & 1.29\\
%     \bottomrule
%   \end{tabular}
%   }
% \end{table}

\begin{table}[t]
  \centering
  \caption{Quantitative comparison in the 64-dim latent space on ImageNet $256\times256$. GAE consistently outperforms existing 64-dim baselines in semantic alignment, reconstruction quality, and generative convergence speed.}
  \label{tab:large_dim}
  \begin{tabular}{lccccccccc}
    \toprule
    \multirow{2}{*}{Method} & \multirow{2}{*}{Dim} & \multirow{2}{*}{LP $\uparrow$} & \multicolumn{4}{c}{Reconstruction} & \multicolumn{3}{c}{gFID (w/o CFG) $\downarrow$} \\
    \cmidrule(lr){4-7} \cmidrule(lr){8-10} % 这里修正为 8-10
    & & & rFID $\downarrow$ & PSNR $\uparrow$ & LPIPS $\downarrow$ & SSIM $\uparrow$ & 80 Ep & 160 Ep & 800 Ep \\
    \midrule
    VTP-L & 64 & 73.9 & \textbf{0.36} & 26.83 & 0.126 & 0.750 & 2.81 & -- & -- \\
    FAE   & 64 & --   & 0.66 & --    & --    & --    & --   & 2.02 & -- \\
    \midrule
    GAE & 64 & \textbf{78.3} & 0.38 & \textbf{29.09} & \textbf{0.081} & \textbf{0.834} & \textbf{1.96} & \textbf{1.53} & \textbf{1.29} \\
    \bottomrule
  \end{tabular}
  \vspace{-1.0em}
\end{table}

\subsection{Ablation Studies}
\subsubsection{Semantic Supervision}
\label{sec:exp_sp_weight}

Our investigation into the sensitivity of the framework to $\lambda_{sp}$ reveals a critical trade-off between reconstruction fidelity and latent discriminability, as detailed in Tab.~\ref{tab:sp_weight_results}. While $\lambda_{sp}=0.5$ yields superior reconstruction metrics, increasing the weight to $\lambda_{sp}=1.0$ provides a more discriminative latent space and higher Linear Probing accuracy with comparable generation quality. This suggests that the marginal loss in reconstruction fidelity is effectively compensated for by the enhanced semantic priors, especially considering that pixel-level quality can be further recovered with extended training duration. However, increasing the weight further to $\lambda_{sp}=2.0$ leads to diminishing returns. The excessive focus on semantic alignment severely compromises reconstruction capacity, creating a bottleneck that results in a suboptimal gFID of 2.45. Consequently, we identify $\lambda_{sp}=1.0$ as the optimal balance that maximizes semantic richness without inducing catastrophic pixel-level degradation.

\begin{table}[t]
\centering
\caption{Ablation of Semantic loss weight $\lambda_{sp}$ on ImageNet-1K. gFID is reported without CFG. Autoencoders are trained on ImageNet-1K for 100 epochs using a consistent 32-dimensional latent bottleneck.}
\label{tab:sp_weight_results}
\begin{tabular}{c|cccc|c|c}
\toprule
$\lambda_{sp}$ & rFID $\downarrow$ & PSNR $\uparrow$ & LPIPS $\downarrow$ & SSIM $\uparrow$ & \textbf{LP} $\uparrow$ & gFID $\downarrow$ \\ \midrule
0.0 & 0.76 & 27.16 & 0.157 & 0.77 & 5.74  & 12.55 \\
0.5 & \textbf{0.40} & \textbf{28.11} & \textbf{0.095} & \textbf{0.796} & 63.5    & \textbf{2.35}  \\
1.0 & 0.45 & 27.37 & 0.107 & 0.774 & 69.2    & 2.36  \\
2.0 & 0.59 & 26.50 & 0.122 & 0.749 & \textbf{71.4}  & 2.45  \\ \bottomrule
\end{tabular}
\vspace{-1.0em}
\end{table}

\subsubsection{The Model Size of Autoencoder}
\label{sec:exp_ae_size}

To investigate the impact of the Autoencoder's parameter, we systematically evaluate three Vision Transformer configurations, including ViT-S, ViT-B, and ViT-L. The quantitative results are summarized in Tab.~\ref{tab:ae_size_results}.

Our analysis reveals that reconstruction fidelity is strongly correlated with the model's representative capacity. While the ViT-L backbone achieves superior performance across all reconstruction metrics, the lightweight ViT-S variant exhibits significant degradation. This scaling also extends to semantic preservation, where ViT-L achieves 69.2\% Linear Probing accuracy, significantly closing the gap toward the 75.6\% performance bound. Notably, despite a slightly higher rFID compared to ViT-B, ViT-L yields a superior gFID of 2.36, reinforcing the notion that enhanced semantic discriminability provides a more effective optimization landscape for latent diffusion.

\begin{wraptable}[8]{r}{0.5\textwidth}
  \centering
  \vspace{-1.5em}
  \caption{Comparison between vanilla VAE and AE with Latent Normalization on ImageNet.}
  \label{tab:rmsnorm_effect}
  % \resizebox{0.48\textwidth}{!}{
  \begin{tabular}{lcc}
    \toprule
    Method & rFID $\downarrow$ & gFID $\downarrow$ \\
    \midrule
    KL constraint & 0.977 & 16.72 \\
    Latent Norm & \textbf{0.764} & \textbf{12.55} \\
    \bottomrule
  \end{tabular}
  % }
  % \vspace{-10pt}
\end{wraptable}

\subsubsection{Latent Normalization}
\label{sec:exp_rmsnorm}
We investigate the impact of normalization strategies by comparing a vanilla VAE against the Autoencoder with Latent Normalization. Both models utilize a ViT-L architecture without sp loss and were trained for 100 epochs. The VAE is trained with a $KL$ weight of $0.1$. As shown in Tab.~\ref{tab:rmsnorm_effect}, while both methods effectively constrain the latent space, the Normalization configuration proves significantly more conducive to diffusion model learning.

\begin{table}[t]
\centering
\caption{Scalability analysis of the Autoencoder backbone. gFID is reported without CFG. Autoencoders are trained on ImageNet-1K for 100 epochs using a consistent 32-dimensional latent bottleneck. }
\label{tab:ae_size_results}
\begin{tabular}{l|ccc|c|cccc|c}
\toprule
Type & Dim & Layers & Heads & LP $\uparrow$ & rFID $\downarrow$ & PSNR $\uparrow$ & LPIPS $\downarrow$ & SSIM $\uparrow$ & gFID $\downarrow$ \\
\midrule
ViT-L & 1024 & 24 & 16 & \textbf{69.2}  & 0.45 & \textbf{27.37} & \textbf{0.107} & \textbf{0.774} & \textbf{2.36} \\
ViT-B & 768  & 12 & 12 & 62.5                & \textbf{0.41} & 26.70          & 0.122          & 0.754          & 2.43 \\
ViT-S & 384  & 12 & 6  & --                  & 1.55          & 23.85          & 0.212          & 0.625          & --   \\
\bottomrule
\end{tabular}
\vspace{-1.0em}
\end{table}

\section{Conclusion}
In this paper, we introduced Geometric Autoencoder (GAE), a principled framework that addresses the fundamental limitations of heuristic latent space design in diffusion models. By systematically analyzing the interplay between semantic alignment, latent distribution, and reconstruction stability, we have moved toward a more grounded architectural foundation. Our approach successfully bridges the gap between high-level perceptual understanding and low-level generative fidelity, ensuring that the latent manifold is both informationally rich and computationally efficient.

GAE demonstrates strong and efficient performance. On ImageNet $256 \times 256$, our model without Classifier Free Guidance achieves a state of the art gFID of 1.31 with 800 training epochs and 1.82 with only 80 epochs. Additionally, GAE establishes a superior equilibrium between compression, semantic depth and robust reconstruction stability. We believe GAE offers a promising roadmap for improving latent diffusion training.

% \clearpage
\bibliographystyle{unsrtnat}
\bibliography{main}

% \newpage

\appendix
\clearpage
\setcounter{page}{1}

\section{Qualitative Results}

\begin{figure}[h]
    \centering
  \includegraphics[width=0.9\linewidth]{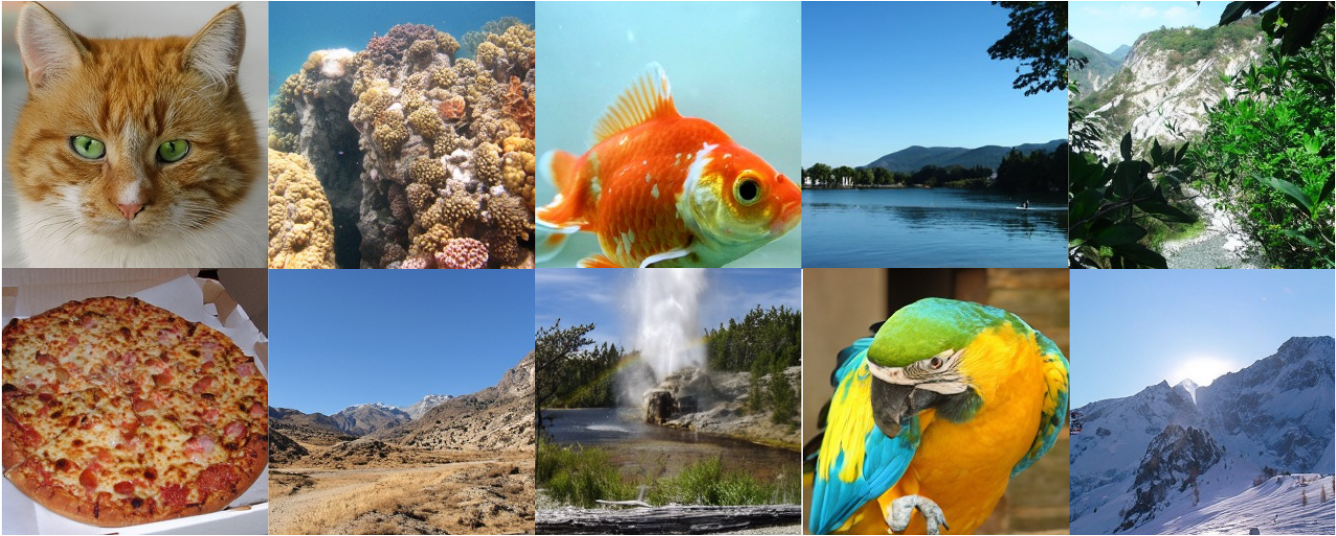}
\caption{\textbf{Qualitative results.} Samples are generated by GAE ($d=32$) after 800 epochs of training. A Classifier-Free Guidance (CFG) scale of $w=3.3$ is utilized to enhance visual fidelity.}
  \label{fig:demo}
\end{figure}

\section{Extended Generative Performance Across Epochs}
\label{sec:supp_gfids}

In this section, Tab.~\ref{tab:gae_gfids_full} provides a detailed breakdown of the generative performance (gFID, without CFG) for GAE with various latent dimensions on the ImageNet $256 \times 256$ benchmark across various training stages. 

\begin{table}[!h]
  \centering
  \caption{Detailed gFID scores without Classifier-Free Guidance for GAE with different latent dimension $d$ at different training epochs on ImageNet $256 \times 256$.}
  \label{tab:gae_gfids_full}
  \begin{tabular}{lcccccc}
    \toprule
    Epochs & 80 & 160 & 320 & 480 & 640 & 800\\
    \midrule
    $d=32$ & 1.82 & 1.51 & 1.37 & 1.33 & 1.31 & 1.31 \\
    $d=64$ & 1.96 & 1.53 & 1.39 & 1.36 & 1.31 & \textbf{1.29} \\
    \bottomrule
  \end{tabular}
\end{table}

\section{Reconstruction Quality}
\label{sec:supp_reconstruction}

In this section, we provide a quantitative comparison of the reconstruction performance between our proposed GAE and various baseline models.

As shown in Tab.~\ref{tab:reconstruction_results}, our model significantly outperforms traditional alignment-heavy frameworks such as RAE and FAE in terms of reconstruction fidelity. Specifically, at a 64-dimensional latent space, our method achieves a PSNR of 29.085 and an SSIM of 0.834, whereas RAE only yields 18.86 PSNR and 0.62 SSIM.

\begin{table}[!h]
  \centering
  \caption{Quantitative comparison of reconstruction quality on ImageNet-1K. We compare our GAE with various generative and alignment-based methods across different latent dimensions ($d$).}
  \label{tab:reconstruction_results}
  \begin{tabular}{lccccc}
    \toprule
    Method & Dim ($d$) & rFID $\downarrow$ & PSNR $\uparrow$ & LPIPS $\downarrow$ & SSIM $\uparrow$ \\
    \midrule
    SD-VAE & 4 & 0.610 & 26.260 & 0.134 & 0.722 \\
    MAR-VAE & 16 & 0.530 & 26.184 & 0.135 & 0.716 \\
    VAVAE & 32 & 0.280 & 27.710 & 0.097 & 0.779 \\
    % VAVAE & 64 & \textbf{0.150} & \textbf{29.130} & \textbf{0.062} & -- \\
    AlignTok & 32 & \textbf{0.260} & 25.830 & 0.117 & -- \\
    % AlignTok & 64 & 0.170 & 27.410 & 0.089 & -- \\
    VTP-L & 64 & 0.360  & 26.830 & 0.126 & 0.750 \\
    RAE (ViT-XL) & 768 & 0.570 & 18.860 & 0.225 & 0.620 \\
    FAE & 32 & 0.680 & -- & -- & -- \\
    FAE & 64 & 0.660 & -- & -- & -- \\
    \midrule
    GAE & 32 & 0.444 & 27.090 & 0.113 & 0.767 \\
    GAE & 64 & 0.382 & \textbf{29.085} & \textbf{0.081} & \textbf{0.834} \\
    \bottomrule
  \end{tabular}
\end{table}

\section{Effect of Semantic Teacher}
\label{sec:exp_sp_teacher}

 We compare two representative architectures: MAE-L, which is optimized for pixel-level reconstruction, and DINOv2-L, which is designed for discriminative representation learning.

As shown in Tab.~\ref{tab:sp_teacher_results}, while MAE-L achieves superior reconstruction metrics (rFID $0.355$, PSNR $28.712$), its limited semantic discriminability ($29.8\%$ LP) leads to a suboptimal gFID of $4.20$. In contrast, DINOv2-L provides significantly stronger semantic priors ($69.2\%$ LP), which, despite a slight decrease in rFID, results in a vastly superior gFID of $2.36$.

% \begin{table}[t]
% \centering
% \caption{Comparison of SP Teachers (MAE-L vs. DINOv2-L) on ImageNet-1K. gFID is reported without CFG.}
% \label{tab:sp_teacher_results}
% \begin{tabular}{l|c|cccc|c|c}
% \hline
% \textbf{Teacher} & \multicolumn{1}{c|}{\textbf{Teacher LP} $\uparrow$} & \multicolumn{4}{c|}{\textbf{AE Reconstruction}} & \multicolumn{1}{c|}{\textbf{AE Latent LP} $\uparrow$} & \textbf{Gen.} \\ \cline{2-8} 
% \textbf{Model} & \textbf{LP}& \textbf{rFID} $\downarrow$ & \textbf{PSNR} $\uparrow$ & \textbf{LPIPS} $\downarrow$ & \textbf{SSIM} $\uparrow$ & \textbf{LP} & \textbf{gFID} $\downarrow$ \\ \hline
% MAE-L   & 33.9  & \textbf{0.36} & \textbf{28.71} & \textbf{0.084} & \textbf{0.813} & 29.8 & 4.20 \\
% DINOv2-L & \textbf{76.0}  & 0.45 & 27.37 & 0.107 & 0.774 & \textbf{69.2} &  \textbf{2.36} \\ \hline
% \end{tabular}
% \end{table}

\begin{table}[!h]
\centering
\caption{Comparison of SP Teachers (MAE-L vs. DINOv2-L) on ImageNet-1K. gFID is reported without CFG.}
\label{tab:sp_teacher_results}
\resizebox{0.9\columnwidth}{!}{
\begin{tabular}{lccccccc}
\toprule
\textbf{Teacher} & \textbf{Teacher LP} $\uparrow$ & \multicolumn{4}{c}{\textbf{AE Reconstruction}} & \textbf{AE Latent LP} $\uparrow$ & \textbf{Gen.} \\ 
\cmidrule(lr){3-6}
\textbf{Model} & \textbf{LP} & \textbf{rFID} $\downarrow$ & \textbf{PSNR} $\uparrow$ & \textbf{LPIPS} $\downarrow$ & \textbf{SSIM} $\uparrow$ & \textbf{LP} & \textbf{gFID} $\downarrow$ \\ 
\midrule
MAE-L    & 33.9 & \textbf{0.36} & \textbf{28.71} & \textbf{0.084} & \textbf{0.813} & 29.8 & 4.20 \\
DINOv2-L & \textbf{76.0} & 0.45 & 27.37 & 0.107 & 0.774 & \textbf{69.2} & \textbf{2.36} \\ 
\bottomrule
\end{tabular}
}
\end{table}

\section{Ablation on CFG Parameters}
\label{sec:supp_cfg_ablation}

In this section, we investigate the impact of various Classifier-Free Guidance parameters on the generative performance of GAE with 32 dimension. Specifically, we examine the CFG weight ($w$), which controls the guidance scale to balance sample fidelity and diversity, and the CFG interval, which defines the temporal range within the sampling process where guidance is actively applied.

\begin{table}[!h]
  \centering
  \caption{Ablation study of CFG parameters for GAE on ImageNet $256 \times 256$. We report the gFID scores across different guidance intervals and weights.}
  \label{tab:cfg_ablation}
  \begin{tabular}{lllll}
    \toprule
    Epochs & Timeshift & CFG Interval & CFG Weight & gFID $\downarrow$ \\
    \midrule
    \multirow{7}{*}{800} & 0.4 & 0.27 & 2.3 & 1.15 \\
                            & 0.4 & 0.27 & 2.5 & 1.14 \\
                            & 0.4 & 0.27 & 3.1 & 1.14 \\
                            & 0.4 & 0.27 & 3.5 & 1.15 \\
                            & 0.4 & 0.3  & 2.7 & 1.15 \\
                            & 0.4 & 0.3  & 3.3 & \textbf{1.13} \\
                            & 0.4 & 0.3  & 3.8 & 1.14 \\
    \midrule
    \multirow{3}{*}{80}  & 0.4 & 0.2  & 2.5 & 1.58 \\
                            & 0.4 & 0.25 & 2.5 & \textbf{1.48} \\
                            & 0.4 & 0.3  & 2.5 & 1.49 \\
    \bottomrule
  \end{tabular}
\end{table}

\section{Algorithm of Patch Convolution}
\label{sec:supp_patch_linear}
As detailed in  Algorithm~\ref{alg:patch_linear}, our patch-wise module is implemented through a two-step procedure designed to optimize computational efficiency. Specifically, we first perform a patch-wise convolution that projects each local spatial block directly into a high-dimensional feature vector. Subsequently, the resulting feature map is re-partitioned and flattened into the target latent patch format.

\begin{algorithm}[ht]
\caption{Patch Convolution with Window Reshaping}
\label{alg:patch_linear}
\begin{algorithmic}[1]
\REQUIRE Feature map $F \in \mathbb{R}^{B \times H \times W \times C}$, window size $w$.
\ENSURE Latent map $Z \in \mathbb{R}^{B \times H \times W \times d}$.

\STATE \COMMENT{Phase 1: Local Window Partitioning}
\STATE $F_{win} \leftarrow \text{WindowPartition}(F, w)$ \COMMENT{Shape: $[B \cdot \frac{HW}{w^2}, w^2, C]$}
\STATE $X \leftarrow \text{Flatten}(F_{win})$ \COMMENT{Flatten each window to $[B \cdot \frac{HW}{w^2}, w^2 \cdot C]$}

\STATE \COMMENT{Phase 2: Patch-wise Joint Projection}
\STATE $Y \leftarrow X W_{proj}$, where $W_{proj} \in \mathbb{R}^{(w^2 \cdot C) \times (w^2 \cdot d)}$ 

\STATE \COMMENT{Phase 3: Spatial Structure Reconstruction}
\STATE $Z_{win} \leftarrow \text{Reshape}(Y, [B \cdot \frac{HW}{w^2}, w^2, d])$
\STATE $Z \leftarrow \text{WindowReverse}(Z_{win}, w)$ \COMMENT{Restore to $[B, H, W, d]$}

\RETURN $Z$
\end{algorithmic}
\end{algorithm}

\begin{figure}[t]
    \centering
  \includegraphics[width=0.6\linewidth]{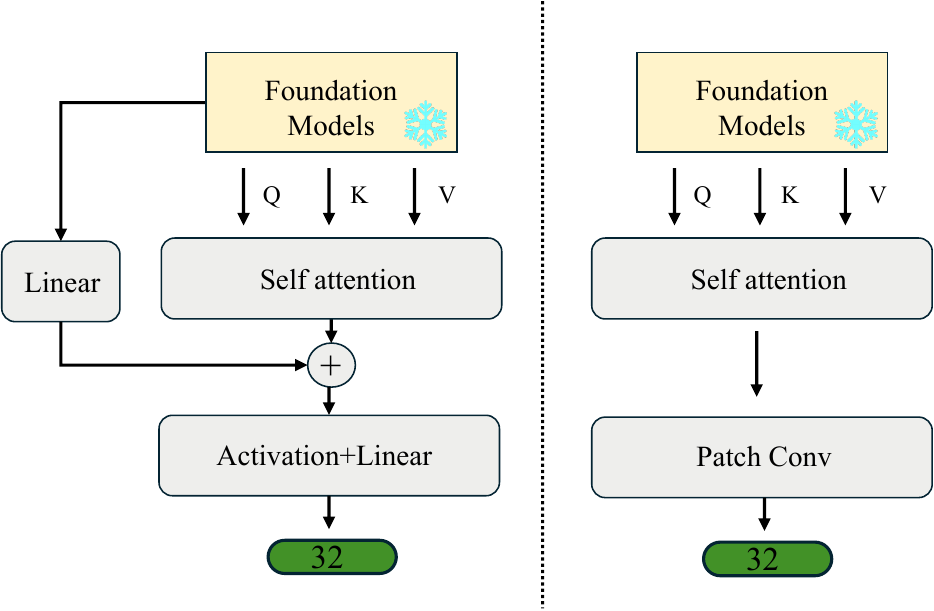}
  \caption{Comparison of downsampler architectures. Left: The FAE downsampler. Right: Our downsampler with patch convolution.}
  \label{fig:sp_encoder_arch}
\end{figure}

\section{Detailed Configuration}
\label{sec:supp_hyp_parameters}
\begin{table*}[ht]
\centering
\caption{Detailed architectural and optimization configurations for main generation. $E_{p}$ and $D_{p}$ denote the pixel encoder and decoder, while $A_{p}$ represents the linear projector. The semantic encoder $E^{*}_{sp}$ integrates a DINOv2-L backbone with downsampler, and $D_{sp}$ serves as an auxiliary decoder for semantic training.}
\resizebox{\textwidth}{!}{
\begin{tabular}{llccccccc}
\toprule
\textbf{Category} & \textbf{Field} &
\textbf{$E_{p}$} & \textbf{$A_{p}$} &\textbf{$D_{p}$} & 
\textbf{$E^{*}_{sp}$} & \textbf{$D_{sp}$} & \textbf{LDM} \\
\midrule

% ---------------- ARCHITECTURE ----------------
\multirow{8}{*}{Architecture}
 & Input dim.        
   & 16$\times$16$\times$3 & 16$\times$16$\times$1024 & 16$\times$16$\times$32
   & 16$\times$16$\times$3 & 16$\times$16$\times$32 & 16$\times$16$\times$32 \\

 & Output dim.       
   & 16$\times$16$\times$1024   & 16$\times$16$\times$32 & 16$\times$16$\times$3
   & 16$\times$16$\times$32 & 16$\times$16$\times$3 & 16$\times$16$\times$32 \\

 & Hidden dim.        
   & 1024 & -- & 1024
   & 1024 & 1024 & 1152 \\

 & Num. layers       
   & 24 & -- & 24
   & 24/1/1 & 4 & 28 \\

 & Num. heads        
   & 16 & -- & 16
   & 16/16/-- & 16 & 16 \\
   
 & MLP Ratio         
   & 4 & -- & 4
   & 4/--/-- & 4 & 4 \\

 & Total Params (M)  
   & 310.38 & 0.03 & 309.36
   & 316.95 & 51.41 & 675.26 \\
\midrule

% ---------------- OPTIMIZATION ----------------
\multirow{7}{*}{Optimization}
 & Training iters   
   & \multicolumn{3}{c}{250k}
   & \multicolumn{2}{c}{10k} & 1M \\

 & Batch size       
   & \multicolumn{3}{c}{1024} & \multicolumn{2}{c}{2048}
    & 1024 \\

 & Optimizer        
   & \multicolumn{3}{c}{AdamW} & \multicolumn{2}{c}{AdamW}
   & AdamW \\

 & Peak LR          
   & \multicolumn{3}{c}{2e-4} & \multicolumn{2}{c}{2e-4}
   & 2e-4 \\

    & Min LR          
   & \multicolumn{3}{c}{2e-5} & \multicolumn{2}{c}{2e-5}
   & 2e-4 \\

 & LR Scheduler     
   & \multicolumn{3}{c}{Cosine} & \multicolumn{2}{c}{Cosine}
    & Constant \\

 & Warmup           
   & \multicolumn{3}{c}{8000} & \multicolumn{2}{c}{500}
   & -- \\

 & $(\beta_1,\beta_2)$
   & \multicolumn{3}{c}{(0.9, 0.98)} & \multicolumn{2}{c}{(0.9, 0.98)}
   & (0.9,0.95) \\

\bottomrule
\end{tabular}
}
\end{table*}

\paragraph{GAE Training Details.}
We instantiate our GAE framework with a ViT-L backbone utilizing a downsampling factor of $f=16$. Semantic supervision is provided by a trained downsampler based on the DINOv2-L architecture. For both 32-dimensional and 64-dimensional latent spaces, we set the semantic preservation loss weight $\lambda_{sp}=1.0$ and the latent standard deviation $C_\sigma=0.2$. The GAE is trained for 200 epochs with a global batch size of 1024. 

\paragraph{Diffusion Model Optimization.}
For the generative process, we strictly adhere to the optimization strategy of LightningDiT-XL/1~\citep{yao2025reconstruction}. We utilize a constant learning rate of $2.0 \times 10^{-4}$, a global batch size of 1024, and an EMA weight of 0.9999. To maintain training stability, QK-Norm is applied for the full 800 epoch training runs, whereas the 80 epoch benchmarks are conducted without QK-Norm to match the short-term evaluation protocol. Furthermore, we incorporate an RAE-style~\citep{zheng2025diffusion} time-shift mechanism to optimize the trajectory of the noise schedule during training.

\paragraph{Sampling and Evaluation Protocol.} 
Following the sampling protocol established in FAE~\citep{gao2025one}, we employ an SDE-based sampler~\citep{song2020score} for evaluations without CFG, and an ODE-based sampler~\citep{chen2018neural} for evaluations involving CFG. All generative results are synthesized using 250 sampling steps. Consistent with RAE~\citep{zheng2025diffusion}, we perform class-uniform sampling during evaluation, ensuring an equal number of samples are generated for each category to provide a balanced assessment of model performance.  For the $d=32$ configuration, hyperparameters are specifically tailored to the training duration: at 800 epochs, we adopt a timeshift of $0.4$, a CFG interval of $0.3$, and a CFG weight of $3.3$. In contrast, for the 80-epoch checkpoint, these parameters are adjusted to $0.4$, $0.25$, and $2.5$, respectively, to optimize for early-stage convergence. For the $d=64$ variant, a timeshift of $0.5$ is consistently employed across our evaluations. To ensure statistical reliability, both gFID and rFID metrics are evaluated using 50,000 images for all experiments.

\section{Detailed Configuration of Ablation Study}
\label{sec:supp_ablation_configs}

Unless specified otherwise in individual experiments, the architectural and optimization settings for our ablation studies follow the general configurations described in Sec.~\ref{sec:supp_hyp_parameters}. 

\paragraph{GAE Training Details.}
For all ablation variants, we instantiate GAE with a ViT-L backbone and a default latent dimension of $d=32$. To accelerate the experimental cycle, GAE is trained for a reduced duration of 100 epochs. Following our noise schedule analysis, the latent standard deviation is set to $C_\sigma=0.1$ for these studies.

\paragraph{Diffusion Model Optimization.}
The diffusion models in the ablation section are trained for 80 epochs, which we found sufficient to reflect relative performance gains. Consistent with our short-term training protocol, QK-Norm is disabled for these runs. All other optimization hyperparameters remain identical to the primary setup defined in Sec.~\ref{sec:supp_hyp_parameters}.

\paragraph{Sampling and Evaluation Protocol.}
Our sampling procedure for ablation follows the protocol established in VAVAE~\citep{yao2025reconstruction}. We primarily report generative performance without classifier-free guidance (CFG) using an ODE-based sampler with 250 steps. Furthermore, class labels are sampled randomly following the VAVAE evaluation standard.
For these ablation experiments, the \textit{timeshift} parameter is fixed at $0.7$. To ensure statistical reliability, both gFID and rFID metrics are evaluated using 50,000 images for all experiments.

\section{Linear Probing Evaluation Protocol}
\label{sec:supp_lp_settings}

To quantitatively evaluate the semantic discriminability of the learned latent space, we report ImageNet-1K linear probing Top-1 accuracy.

Unlike prior works that rely on DINO-style Linear Probing~\citep{gao2025one,yao2025towards} by concatenating pooled latents with the original DINO CLS tokens, we assess the learned latents as stand-alone representations to strictly measure their inherent semantic quality. Specifically, we conduct LP tasks using two distinct configurations based on the feature dimensionality in the main paper. We also provide the full set of LP results in Tab.~\ref{tab:compre_lp_result}:

\begin{itemize}
    \item Flatten: For compressed latents (e.g., $d=32$ or $64$), we map the spatial tensor $\mathbb{R}^{d \times 16 \times 16}$ directly to the 1,000 ImageNet classes. This captures both spatial and semantic information within the bottleneck.
    \item GAP: For high-dimensional, uncompressed representations (e.g., $1,024$), we apply Global Average Pooling to collapse the spatial dimensions into a vector $\mathbb{R}^{1024}$ prior to classification. This prevents the excessive parameter count associated with flattening high-dimensional tensors, which otherwise hinders linear optimization.
\end{itemize}

\paragraph{Optimization Details.}
The linear probes are trained for $100$ epochs using a global batch size of $2,048$. We utilize a Stochastic Gradient Descent (SGD) optimizer with a constant learning rate of $0.008$. No weight decay or complex learning rate scheduling is applied during this phase.

\begin{table}[h]
\centering
% \vspace{-1.0em}
\caption{Comprehensive LP results.}
% \vspace{-1.0em}
\label{tab:compre_lp_result}
\begin{tabular}{lccc} % 去掉了所有竖线 |
\toprule % 最顶部的粗线
\textbf{Method} & \textbf{Dim} & \textbf{Flatten} $\uparrow$ & \textbf{GAP} $\uparrow$ \\ 
\midrule % 标题下的分隔线

\multicolumn{4}{l}{\textit{Part 1: Linear Probing results for Fig.~\ref{fig:trilemma}}} \\ \midrule

SD-VAE &16 & 4.99 & 1.16 \\
MAR-VAE &16 & 13.7 & 4.29 \\
VAVAE & 32 & 43.1 & 29.0 \\
Aligntok & 32 & -- & 35.1 \\
VTP-L & 64 & 73.9 & 67.6 \\
RAE & 768 & 74.8 & \textbf{81.7} \\
GAE & 32 & 69.4 & 43.9 \\
GAE & 64 & \textbf{78.3} & 56.8 \\

\midrule % 各部分之间用 midrule 分隔
\multicolumn{4}{l}{\textit{Part 2: Linear Probing results for Tab.~\ref{tab:pre_mid_post}}} \\ \midrule

Pre Alignment             & 32                    & 20.9                      & 9.91                   \\
Post Alignment            & 32                    & 60.8                      & 44.7                   \\
Latent Alignment            & 32                    & \textbf{63.2}             & \textbf{47.5}          \\
Teacher (SVD)    & 32                    & 60.9                      & 48.9                   \\
Teacher (DinoL)  & 1024                  & 77.5                      & 83.7                   \\ 

\midrule
\multicolumn{4}{l}{\textit{Part 3: Linear Probing results for Tab.~\ref{tab:sp_ablation}}} \\ \midrule

Single Attn            & 32  & 62.8        & 51.0       \\
Attn + Linear     & 32  & 63.4        & \textbf{52.3}      \\
Attn + Patch Conv & 32  & \textbf{75.6} & 48.9 \\ 

\midrule
\multicolumn{4}{l}{\textit{Part 4: Linear Probing results for Tab.~\ref{tab:sp_weight_results}}} \\ \midrule

$\lambda_{sp} = 0$ &32 &5.74 &2.15 \\
$\lambda_{sp} = 0.5$ &32 &63.5 &42.7 \\
$\lambda_{sp} = 1.0$ &32 &69.2 &43.9 \\
$\lambda_{sp} = 2.0$ &32 &\textbf{71.4} &\textbf{44.3} \\

\midrule
\multicolumn{4}{l}{\textit{Part 5: Linear Probing results for Tab.~\ref{tab:ae_size_results}}} \\ \midrule

ViT-L &32 & \textbf{69.2} &\textbf{43.9} \\
ViT-B &32 &62.5 & 41.1 \\

\midrule
\multicolumn{4}{l}{\textit{Part 6: Linear Probing results for Tab.~\ref{tab:sp_teacher_results}}} \\ \midrule

MAE-L (AE) &32 & 29.8 & 13.5 \\
DINOv2-L (AE) &32 & 69.2 & 43.9 \\
Teacher (MAE-L) &32 &33.9 &18.3 \\
Teacher (DINOv2-L) &32 & 76.0 & 49.0 \\
\bottomrule % 最底部的粗线
\end{tabular}
\vspace{-5.0em}
\end{table}

\end{document}